\documentclass[a4paper]{article}

\usepackage{INTERSPEECH2019}
\usepackage{xcolor,amsmath,amsfonts,url}
\usepackage{hyperref}
\def\R{\mathbb{R}}

\title{Unsupervised Singing Voice Conversion}
\name{Eliya Nachmani$^{1,2}$, Lior Wolf$^{1,2}$}
\address{
  $^1$Facebook AI Research\\
  $^2$The School of Computer Science, Tel Aviv University}
\email{enk100@gmail.com,wolf@cs.tau.ac.il}

\begin{document}

\maketitle
\begin{abstract}
  We present a deep learning method for singing voice conversion. The proposed network is not conditioned on the text or on the notes, and it directly converts the audio of one singer to the voice of another. Training is performed without any form of supervision: no lyrics or any kind of phonetic features, no notes, and no matching samples between singers. The proposed network employs a single CNN encoder for all singers, a single WaveNet decoder, and a  classifier that enforces the latent representation to be singer-agnostic. Each singer is represented by one embedding vector, which the decoder is conditioned on. In order to deal with relatively small datasets, we propose a new data augmentation scheme, as well as new training losses and protocols that are based on backtranslation. Our evaluation presents evidence that the conversion produces natural signing voices that are highly recognizable as the target singer. 
\end{abstract}

\section{Introduction}

The singing human voice is arguably the most important musical instrument in existence. Recently, deep neural networks have been applied successfully for synthesizing singing voices, based on the language features and the accompanying notes. In this work, we explore a related application, in which a singer's voice is converted to another singer's voice. This could lead, for example, to the ability to free oneself from some of the limitations of one's own voice. While existing pitch correction methods, such as the Auto-Tune software, correct local pitch shifts, our work offers flexibility along the other voice characteristics. 

Our method is unsupervised and does not employ supervision of any form. We do not require parallel training data between the various singers, nor do we employ a transcript of the audio to either text (i.e., phonemes) or to musical notes. This makes the sample collection process much simpler than what is required for most literature singer conversion methods. Singing samples are abundant, and, if needed, the technology for separating voice from instrumental music has advanced greatly, due to the advent of deep learning.

From a technical perspective, our work presents multiple technical novelties, including a new training scheme that combines backtranslation and the mixup technique, and new data augmentation techniques. Together, we are able to learn to convert between singers from 5-30 minutes of their singing voices. Our contributions are: (i) the first method, as far as we know, to perform an unsupervised singing voice conversion, where the target singer is modeled from a different song, (ii) demonstrating the effectiveness of a single encoder and a conditional decoder trained in an unsupervised way, (iii) introducing a two-phase training approach in unsupervised audio translation, in which backtranslation is used in the second phase, (iv) introducing backtranslation of mixup (virtual) identities, (v) suggesting a new augmentation scheme for training in a data efficient way.

\section{Related work}

Our method is based on a WaveNet~\cite{wavenet} autoencoder. These autoencoders were used to model single musical instruments~\cite{nsynth2017} and extended to perform translation between musical domains in~\cite{mor2018autoencoderbased} by employing a single encoder and multiple decoders. The translation is done without parallel data in a method that is similar to our first phase of training, except that we employ a single, singer conditioned, WaveNet decoder and a different data augmentation procedure. Most previous work that employ a WaveNet decoder that is conditioned on the embedding of the speaker~\cite{deepvoice3,tacotron2,chen2018sample}, employ supervised learning, while we employ unsupervised learning.

In the unsupervised VQ-VAE method~\cite{NIPS2017_7210}, voice conversion was obtained by employing a WaveNet autoencoder that produces a quantized latent space. The decoder is conditioned on the target speaker's identity, using a one-hot encoding. The strong bottleneck effect achieved by discretization, leads to an embedding that is supposedly speaker-invariant. In our work, following~\cite{mor2018autoencoderbased,polyak2019attention}, we employ a domain confusion loss. As shown in~\cite{mor2018autoencoderbased}, for the task of voice conversion, the domain confusion approach outperforms the discrete code approach of~\cite{NIPS2017_7210}.

Other autoencoder-based approaches in the field of voice conversion have relied on variational auto encoders~\cite{vae} to generate spectral frames. In~\cite{hsu2016voice}, the notion of a single encoder and a parameterized decoder, where the parameters represent the identity, was introduced. The method was subsequently improved~\cite{hsu2017voice} to include a WGAN~\cite{wgan} to improve the naturalness of the output (not as a domain confusion term). 

\noindent{\bf Singing Synthesis and Conversion\quad} Classical singing synthesis methods are mostly concatenative (unit selection) methods~\cite{bonada2016expressive} or HMM based~\cite{saino2006hmm,oura2010recent}. Blaaauw and Bonda have demonstrated very convincing singing synthesis using a WaveNet Decoder~\cite{singing}.  Their system receives, as input, both notes and lyrics and produces a stream of vocoder features. The method was extended~\cite{singing2} to adopt between singers, using the same type of note and lyrics supervision, in a data efficient manner, based on a few minutes of clean audio per target singer.

In the field of singing voice conversion, i.e., transforming an audio of a song to a target voice, almost all literature methods have used parallel data~\cite{kobayashi2015statistical,kobayashi2014statistical,conf/interspeech/VillavicencioB10}, i.e., different singers that are required to perform the same song. None of these existing methods provide code or benchmarks that can be used for a direct comparison of their results (even if supervised) to ours.

Very recently, a method that does not require parallel data was presented, in which the acoustic features of the target singer are extracted from their speech (not from a song)~\cite{snap}. Vocoder features are used for synthesizing the audio. The results are demonstrated on four source singers, one target voice, and as can be heard in their sample page~\href{https://sites.google.com/site/singingvoiceconversion2018/} are still partly convincing. 

\noindent{\bf Backtranslation\quad} The technique of back-translation has emerged in Natural Language Processing, where it was presented as a technique for employing monolingual corpora in the supervised training of an automatic machine translation (AMT) system~\cite{sennrich2015improving}. A sample $a$ in language $A$, which does not have a matching translation in the target language $B$, is automatically translated by the current AMT system to a sample $b$ in that language. One then considers the training pair ($b$,$a$) for translating from the language $B$ back to language $A$. 

Since our conversion system is symmetric (it can convert in both directions), we can backtranslate. However, after the first training phase of our method, the conversion offered by the network is still very limited. Synthetic samples are, therefore, created using virtual identities that are closer to the source singer than other speakers.

\noindent{\bf Mixup training\quad} The mixup techniques for learning a function $y=f(x)$ trains on an additional set of virtual samples $(x',y')$ created by combining two samples $(x_1,y_1)$ and $(x_2,y_2)$ using the same random weight $\beta \in [0,1]$ that is sampled from the Beta distribution  $x' = \beta x_1 + (1-\beta) x_2$ and $y' = \beta y_1 + (1-\beta) y_2$. In the literature, the shape parameters of the Beta distribution are taken to be 0.2, which results in sampled values that are near one of the edges, i.e., a $\beta$ that is often close to zero or to one. 

In our application of mixup, there are two modifications: first, since we employ backtranslation, we do not require the generation of a mixed audio, just the generation of a conversion to a mixed identity. Second, since we are interested in samples that are not concentrated away from the first sample, we replace the beta distribution with a uniform one.

\begin{figure}[t]
  \centering
\includegraphics[width=0.95\linewidth,trim={0 0 0 0},clip]{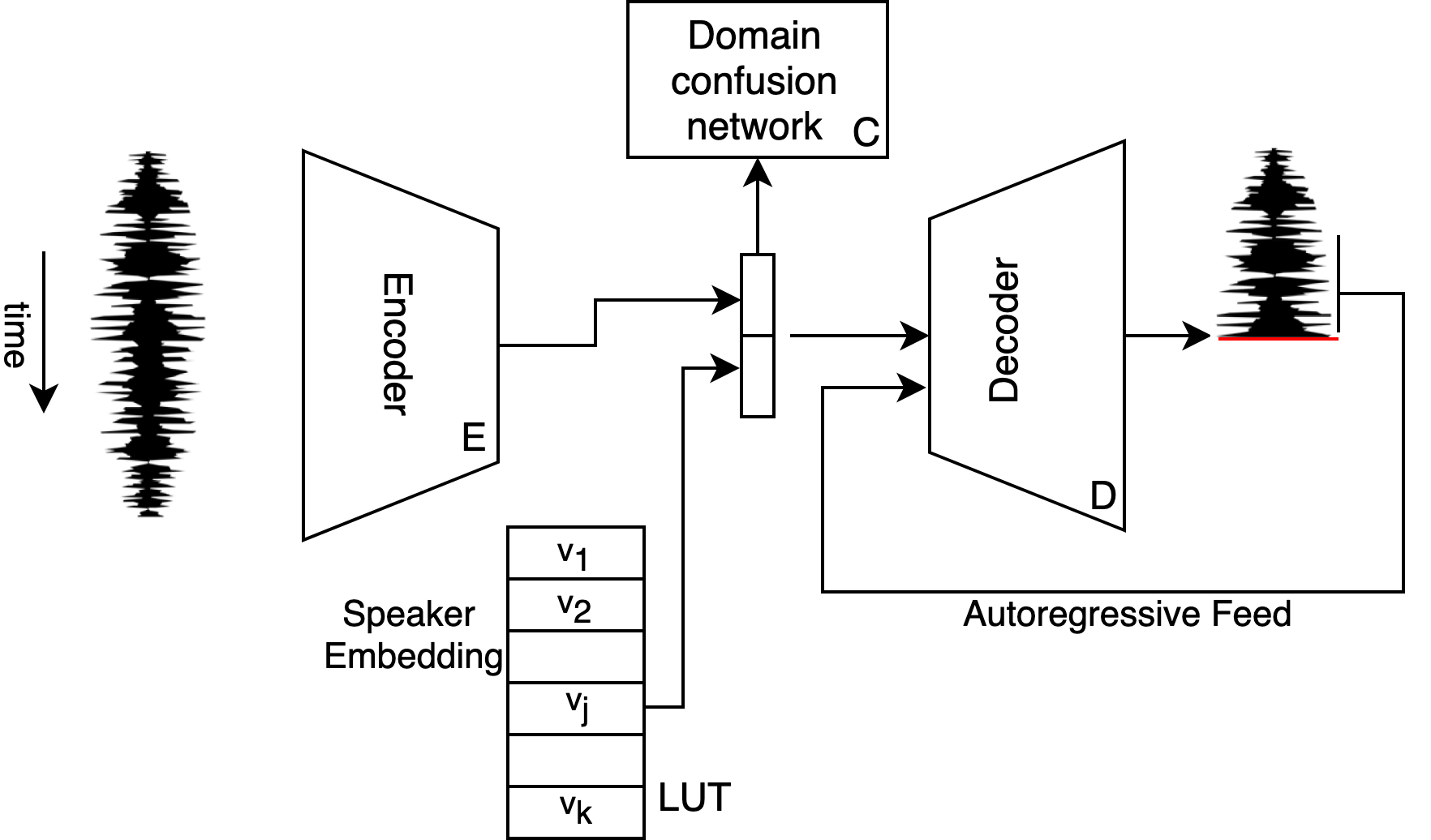} \\
  \caption{The schematic architecture of our singing voice conversion network. We employ an encoder $E$, a domain confusion network $C$ and a conditional decoder $D$. The speakers' embeddings $v_j$ are stored in a LUT. The conditioning of $D$ is on a concatenation of the speaker's embedding and the output of the encoder at every time point.}
 \label{fig:arch}
\end{figure}

\begin{figure}[t]
  \centering
 \begin{tabular}{c}
 \includegraphics[width=.9\linewidth,clip]{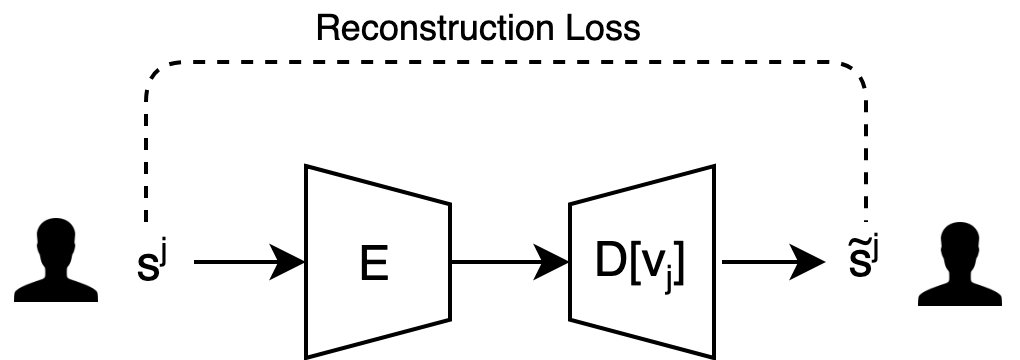} \\
 (a) \\
 ~\\
 \includegraphics[width=.9\linewidth,clip]{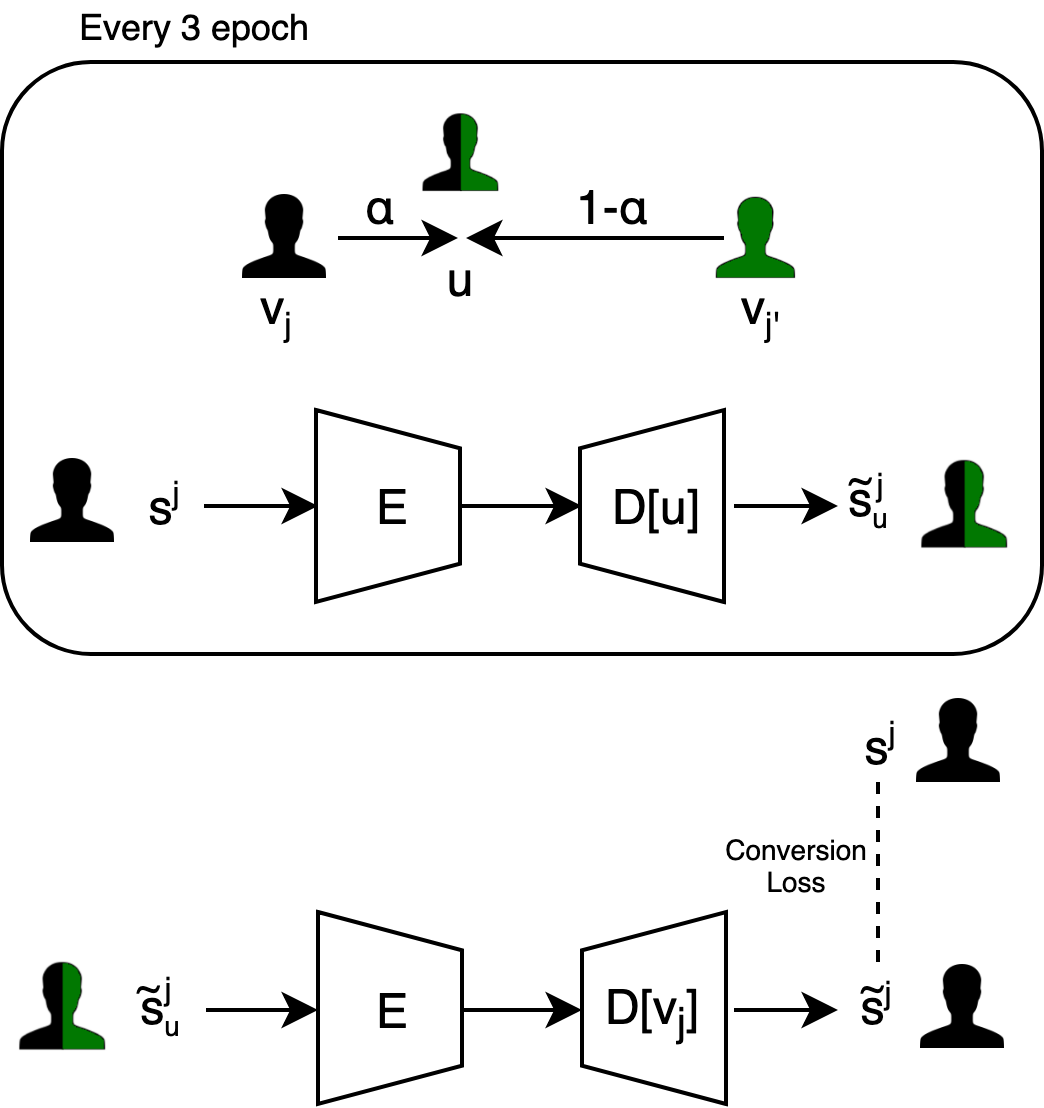} \\
 (b)\\
 \end{tabular}
  \caption{The two training phases. (a) Phase I in which only the reconstruction loss (shown) and the domain confusion loss (not show) are applied. Samples are being reconstructed using the autoencoding path of the same singer $j$. (b) In the second phase of training, synthetic samples are being generated by converting from a sample of singer $j$ to a mixup voice that combines the vocal characteristics of two singers $j$ and $j'$. These samples are used as training samples for a backtranslation procedure, in which they are translated back to singer $j$ and a loss then compares to the original sample of singer $j$.}\label{fig:arch2}
\end{figure}

\section{Method}

The singing voice conversion method employs a single encoder and a singer decoder, which is conditioned on a vector embedding of the target singer.  
There are two phases of training. In the first phase, a softmax-based reconstruction loss is applied to the samples of each singer separately. In the second, samples of novel singers obtained by mixing the vector embeddings of the training singers are created and the network is also trained for successfully converting these synthetic samples back to the training sample used to create them. In addition to the reconstruction losses, a domain confusion loss~\cite{Ganin:2016:DTN:2946645.2946704} ensures that the latent space of the encoder is singer-agnostic. 

\subsection{The Conversion Network}

A diagram of the proposed architecture is shown in Fig.~\ref{fig:arch}. Let $s$ be an input sample from any singer and $s^j$ be an input sample from singer $j=1,2,\dots,k$, $k$ being the number of singers in the training set. The sample can be an original sample or one generated by the augmentation process of Sec.~\ref{sec:aug}. Let $E$ be the encoder, and $D[u]$ be the WaveNet decoder conditioned on the vector $u$. Let $C$ be the singer classification network. Finally, let $v_j \in \R^{64}$ be the learned vector embedding of singer $j$. A Look Up Table (LUT) stores the embedding vector for each of the singers. At each training iteration, each embedding vector with a norm larger than $1.0$ was normalized to have a norm of $1.0$, making all embedding vectors lie within the unit sphere. 

The confusion  network $C$ predicts the singer associated with the input sample $s$ based on the latent vectors $E(s)$ generated by the encoder. Training is done in two phases. During the first phase of training, illustrated in Fig.~\ref{fig:arch2}(a), the confusion network $C$ minimizes the classification loss
\begin{equation}
\sum_j \sum_{s^j} \mathcal{L}(C(E(s^j)),j),
\end{equation}
and the $k$ autoencoder pathways $D[v_j]\circ E$, $j=1,2,\dots,k$ are trained with the loss
\begin{equation}
 \sum_j \sum_{s^j}  \mathcal{L}(D[v_j](E(s^j)),s^j) -\lambda\sum_j \sum_{s^j} \mathcal{L}(C(E(s^j)),j)
 \label{eq:lossphase1}
\end{equation}
where $\mathcal{L}(o,y)$ is the cross entropy loss applied to each element of the output $o$ and the corresponding element of the target $y$ separately, and $\lambda$ is a weight factor. The decoders $D[v_j]$ are autoregressive models, which are conditioned on both the singer embedding vector $v_j$ and the output of $E$. During training, the autoregressive model is fed the target output $s^j$ from the previous time-step, instead of the generated output, i.e., it is trained using what is often called ``teacher forcing''. 

The network that is trained with the loss of of Eq.~\ref{eq:lossphase1} is able to reconstruct the original signal, and its encoder produces an embedding that is (somewhat) singer-agnostic. However, it is not trained directly to perform a singer translation. Therefore, it shows only a limited success in this task.

The second phase of training is illustrated in Fig.~\ref{fig:arch}(b). In this phase, backtranslation is applied in order to create parallel samples and train the network on these samples. This is done in combination with the mixup technique~\cite{zhang2017mixup} in order to generate ``in-between'' singers that are easier to fit.

Every mixup sample $s^j_u$ is based on a mixup singer embedding $u$, which is constructed based on the embedding of two different singers $j$ and $j'$, at some point during training, as:
\begin{equation}
    u = \alpha v_j + (1-\alpha) v_{j'},
\end{equation}
where $\alpha\sim U[0,1]$ is drawn from the uniform distribution. The sample $s^j_u$ is generated during training by transforming a sample $s^j$ of singer $j$ with the current network:
\begin{equation}
    s^j_u = D[u](E(s^j)
    \label{eq:mixupsample}
\end{equation}

Once the second phase of training starts, we create new mixup samples every three epochs and use them in order to add the following loss to the training of $D$ and $E$ (we do not train $C$ with these):
\begin{equation}
   \sum_{s^j_u}  \mathcal{L}(D[v_j](E(s^j_u)),s^j),
\end{equation}
where $s^j$ is the audio clip used to generate $s^j_u$ in Eq.~\ref{eq:mixupsample}.

At test time, in order to convert a sample $s$ in any singer's voice to the voice of singer $j$, we apply the conditioned autoencoder pathway of domain $j$, obtaining the new sample $D[v_j](E(s))$. The bottleneck during inference is the autoregressive decoding process, which is performed in real time,  using the the dedicated CUDA kernels described in~\cite{mor2018autoencoderbased}.

\subsection{Audio Input Augmentation}
\label{sec:aug}
\begin{table*}[ht]
\begin{minipage}[c]{0.49\linewidth}
\caption{MOS Naturalness Scores (Mean $\pm$ SE). Higher is better.\label{tab:mos_nat}}
\begin{tabular}{lccc}
\toprule
 & Ground truth & Reconstruction & Conversion    \\
\midrule
DAMP		 & 4.30$\pm$0.90 & 4.01$\pm$0.91 & 3.99$\pm$0.83 \\
NUS (male) 	 & 4.31$\pm$0.83 & 4.12$\pm$0.86 & 4.10$\pm$0.72 \\
NUS (female) & 4.34$\pm$0.73 & 4.01$\pm$0.73 & $4.00\pm$0.98 \\
NUS (all)    & 4.37$\pm$0.82 & 4.14$\pm$0.79 & 4.09$\pm$0.87 \\
\bottomrule
\end{tabular}
\smallskip
\smallskip
\smallskip
\smallskip
\smallskip
\end{minipage}%
\hfill
\begin{minipage}[c]{0.49\linewidth}
\centering
\caption{MOS Similarity Scores (Mean $\pm$ SE)\label{tab:mos_sim}}
\begin{tabular}{lcc}
\toprule
 & Reconstruction & Conversion    \\
\midrule
DAMP		& 4.20$\pm$0.91 & 3.01$\pm$1.31 \\
NUS (male) 	& 4.54$\pm$0.70 & 3.69$\pm$1.13 \\
NUS (female) & 4.35$\pm$0.93 & 3.54$\pm$1.21 \\
NUS (all)    & 4.59$\pm$0.64 & 3.44$\pm$1.11 \\
\bottomrule
\end{tabular}
\smallskip
\smallskip
\smallskip
\smallskip
\smallskip
\end{minipage}
\begin{minipage}[c]{0.39\linewidth}
\caption{ Identification Accuracy Top1\label{tab:identification}}
\begin{tabular}{lcc}
\toprule
Dataset & Reconstruction & Conversion    \\
\midrule
DAMP		& 93.20$\%$ & 91.60$\%$ \\
NUS (male) 	& 100$\%$ & 96.15$\%$ \\
NUS (female) & 100$\%$ & 92.30$\%$ \\
NUS (all)    & 100$\%$ & 95.00$\%$ \\
\bottomrule
\end{tabular}
\end{minipage}%
\hfill
\begin{minipage}[c]{0.62\linewidth}
\caption{ The identification accuracy for multiple variants of our method\label{tab:ablation}}
\begin{tabular}{lccc@{}}
\toprule
 & NUS (male) & NUS (female) & NUS (all)    \\
\midrule
Full method & 96.15$\%$ & 92.30$\%$ & 95.00$\%$\\
Phase II without mixup & 84.61$\%$ & 92.30$\%$ & 93.33$\%$\\
Phase I only (no backtranslation)& 76.92$\%$ & 80.76$\%$ & 71.66$\%$\\
Phase I no augmentation & 65.38$\%$ & 53.84$\%$ & 68.33$\%$ \\
\bottomrule
\end{tabular}
\end{minipage}
\end{table*}

The audio in the network, including both input and output, follows an 8-bit mu-law encoding, as in previous work~\cite{wavenet,nsynth2017}. This bounds the quality of the audio produced by the network, but supports efficient training in a straightforward way.

In the music translation network of~\cite{mor2018autoencoderbased}, the pitch of the input audio clip was changed locally, in order to enforce the encoder to maintain semantic information and not memorize the input signal. We found that this augmentation is detrimental for our task and, therefore, do not employ such an augmentation. 

A major obstacle for training our method is the limited amount of training data in the current datasets, which consist of 4 or 9 relatively short songs per singer. Therefore, the main goal of our augmentation scheme is to generate more data. 

In order to perform the augmentation, we make use of the well-known fact that when a signal is played backward, the energy spectrum does not change~\cite{hartmann2004signals}. Another fact that we rely on, is that the audio signal presents the change of pressure amplitude at a certain point in space, known as acoustic wave. The human perception for monaural audio is not affected by the phase of the signal, hence one can shift the phase by 180 degrees (multiply by -1)  without any effect on auditory perception. 

The augmentation method we propose increases by four fold the size of the dataset. This is done by first playing each song both forward and backward in time, and second, by multiplying the values of the raw audio signal by -1. The first augmentation creates a  gibberish song that is nevertheless identifiable as the same singer; the second augmentation creates a perceptually indistinguishable but novel signal for training.

\subsection{The Architecture of the Sub-Networks} \label{wavenet_autoencoder}

The autoencoder network consists of a WaveNet-like dilated convolution encoder $E$ and on a WaveNet decoder $D$. The decoder is conditioned on the latent representation produced by the encoder and on a vector embedding of the singer. The architecture of the encoder, decoder, and confusion network mostly reuse the successful WaveNet autoencoder architecture~\cite{nsynth2017,mor2018autoencoderbased}. 

The encoder $E$ is a fully convolutional network, which contains three blocks of ten residual-layers, a total of thirty layers. Each residual-layer comprises of a RELU nonlinearity, a non-causal dilated convolution, a second RELU, and a $1\times 1$ convolution followed by the residual summation of the activations before the first RELU. A fixed width of 128 channels is employed. After the three blocks,  an additional $1\times 1$ layer passes the data to an average pooling layer, with a kernel size of 50 milliseconds (800 samples). The resulting encoding in $\mathbb{R}^{64}$,  has a temporal down sampling  factor of $\times 12.5$.

In order to condition the WaveNet decoder, the audio encoding, given by the encoder, is concatenated to the target singer embedding $v_j$, resulting in a vector of dimensinality 128. The first half of this vector varies in time, while the second part is fixed as long as the singer does not change. The encoding is then upsampled temporally to the original audio rate.

The conditioning signal is passed through a $1\times 1$ layer and fed multiple times to the WaveNet decoder, where each residual layer of the WaveNet receives the conditioning signal after a different $1\times 1$ layer. The WaveNet decoder has four blocks of 10 residual-layers as in~\cite{nsynth2017}, leading to a receptive field of 250 milliseconds (4,093 raw samples). Each residual-layer contains a causal dilated convolution with an increasing kernel size, a gated $tanh$ activation, a $1\times 1$ convolution followed by the residual summation of the layer input, and a $1\times 1$ convolution layer which introduces a skip connection. The summed skip connections, together with the conditioning signal, are passed through two fully connected layers and a softmax activation to output the probability of the signal in the next timestep. 

The architecture of the confusion network $C$ follows the architecture of~\cite{mor2018autoencoderbased}, and applies three 1D-convolution layers, with the ELU~\cite{elu} nonlinearity. The last layer projects the vectors to dimension $k$ (the number of singers) and the vectors are subsequently averaged to a single $\R^k$ vector.

\section{Experiments}

Since we are unaware of public implementations of singing voice conversion systems (supervised or unsupervised), nor of suitable public benchmarks, we focus our empirical validation on a comparison to the ground truth songs. 

We employed two publicly available datasets. The first is the Smule DAMP Dataset~\cite{smith2013correlation,wang2018singing} \url{https://ccrma.stanford.edu/damp}, accessed April 1, 2019. From the ``DAMP-multiple'' section of this dataset, we selected five singers at random (ID 84955650, 84032291, 102087231, 101442956, 100881623), after excluding several singers with low quality audio. Each singer has $10$ vocal songs, out of which $9$ songs are used for training, and the tenth for validation. 

The second dataset is NUS-48E \cite{duan2013nus}. The dataset contains $12$ singers with four songs for each singer, all of which are used for training our unsupervised method, disregarding the speech files from the dataset. The audio in this dataset is significantly cleaner than in the DAMP dataset and we train three networks with it: one network for the 12 singers, one network for the six male singers, and one for the six female singers.

When evaluating the results, we use both automatic and user-study based success metrics: (i) Mean Opinion Scores (MOS) of the quality, on a scale between 1--5. These were computed using the ``same\_sentence'' option of crowdMOS, following~\cite{deepvoice2} (personal communication). (ii) MOS of the similarity of the generated voice to the target singing voice, on a scale between 1--5. 
(iii) Automatic identification by employing a CNN trained for speaker recognition.

Sample results are shared in our supplementary material and online~\url{https://enk100.github.io/Unsupervised_Singing_Voice_Conversion/}.  Tab.~\ref{tab:mos_nat} presents MOS scores for quality. As can be seen, both the reconstructed (no translation) audio and the converted audio present scores of about 4, which is considered good quality. Interestingly, the converted results are not significantly lower in quality than the reconstructed ones.

The MOS scores for similarity are shown in Tab.~\ref{tab:mos_sim}. As can be seen, the identity similarity of the converted music is relatively high (3.0--3.7) and is higher in the NUS dataset in comparison to the noisier DAMP dataset. 

The similarity of the generated voices to the target voices is also evaluated automatically, using the same type of classification network that was used to verify voice identifiability in previous work~\cite{deepvoice2,taigman2017voice,nachmani2018fitting,nachmani2019unsupervised}. A multi-class CNN is trained on the ground-truth training set of the multiple singers, and tested on the generated samples. The network employs the world vocoder features~\cite{worldvocoderfeatures}, stacked across the time domain to form an ``image''. Its architecture consists of five convolutional layers of 3$\times$3 filters with 32 batch-normalized and ReLU activated channels, followed by max-pooling, average pooling over time, and two fully-connected layers. The final layer has a softmax output as large as the number of training singers.

Since the NUS dataset is small (4 songs per singer), we train the identification network with the entire the dataset. Note, however, that we are testing identification on the converted samples.  For DAMP dataset, we use the same training-validation split as used to train the proposed architecture. 
The identification results are shown in Tab.~\ref{tab:identification}. As shown, the identification accuracy of the generated samples is almost as high as those of the reconstructed samples. 

An ablation analysis to highlight the contribution of the various components of our system. The major challenge is not to reconstruct relatively high quality audio but to perform conversion that results in the target speaker. Therefore, our ablation analysis focuses on identification, which is also easier to test automatically. As can be seen from Tab.~\ref{tab:ablation}, mixup provides an advantage in comparison to backtranslating between dataset speakers $j$ and $j'$ (the phase II no mixup row). When eliminating the 2nd phase altogether, i.e., continuing to train without backtranslation, performance greatly suffers. We also observe a significant gap in the accuracy when training the conversion model without the data augmentation (tested only for phase I).

\section{Conclusions}


Our unsupervised method is shown to produce high quality audio, which is recognizable as the target voice. The method is based on a single CNN encoder and on a single conditional WaveNet decoder. A confusion network encourages the latest space to be singer-agnostic and new types of augmentation are applied in order to overcome the limited amount of available data. Crucially, training is done in two phases, by employing a backtranslation method that employs mixup identities.

As future work, we would like to find out whether a similar method can perform the conversion in the presence of background music. We believe that this can be done in an unsupervised way, without relying on a supervised voice separation technique for preprocessing.


\bibliographystyle{IEEEtran}
\bibliography{voice}

\end{document}